# Textile Pattern Generation Using Diffusion Models

**Halil Faruk Karagoz[1,*], Gulcin Baykal[1], Irem Arikan Eksi[2], Gozde Unal[3]**

[1] Istanbul Technical University, Computer Engineering Department, Istanbul, Turkey
[2] Bilgi University, Textile and Fashion Design Department, Istanbul, Turkey
[3] Istanbul Technical University, AI and Data Engineering Department, Istanbul, Turkey

[*] *karagozh18@itu.edu.tr*

**ABSTRACT**

The problem of text-guided image generation is a complex task in Computer Vision, with various applications, including creating visually appealing artwork and realistic product images. One popular solution widely used for this task is the diffusion model, a generative model that generates images through an iterative process. Although diffusion models have demonstrated promising results for various image generation tasks, they may only sometimes produce satisfactory results when applied to more specific domains, such as the generation of textile patterns based on text guidance. This study presents a fine-tuned diffusion model specifically trained for textile pattern generation by text guidance to address this issue. The study involves the collection of various textile pattern images and their captioning with the help of another AI model. The fine-tuned diffusion model is trained with this newly created dataset, and its results are compared with the baseline models visually and numerically. The results demonstrate that the proposed fine-tuned diffusion model outperforms the baseline models in terms of pattern quality and efficiency in textile pattern generation by text guidance. This study presents a promising solution to the problem of text-guided textile pattern generation and has the potential to simplify the design process within the textile industry.

## 1. INTRODUCTION

Textile pattern generation, which involves creating high-quality patterns based on text guidance for industries such as fashion, home decor, and industrial design, is challenging due to the diverse range of patterns and styles and the complex relationship between patterns and text descriptions. However, synthetic image generation, which is the artificial creation of images that resemble natural images, has emerged as a popular area of research in AI and has the potential to significantly simplify the textile pattern design process. Recently, there has been a shift in the field of generative AI, with Denoising Diffusion models [1] emerging as the new state-of-the-art technology. This change is particularly evident in image generation, where these models have outperformed Generative Adversarial Networks (GANs) [2], which have been the dominant technology for a significant time.

On the other hand, the CLIP [3] model is a highly utilized AI model enabling text-guided image generation by connecting visual representations with natural language supervision. This hybrid model has been widely adopted for text-guided image generation, allowing for creating images based on text descriptions.

We choose the Stable Diffusion model [4] for this work due to its open-source availability and proven ability to produce high-quality images for various tasks. The use of diffusion models [1] in image generation has significantly advanced the field, allowing for creating realistic images for various purposes.

Our goal is to produce pattern images for textiles using the Stable Diffusion model and a given prompt text. Although Stable Diffusion effectively generates scenes, it falls short in generating patterns for textiles. To overcome this limitation, we fine-tune the model using custom datasets tailored for textile pattern generation. The pattern images are collected in various styles, and the images are then captioned using the BLIP [5] model. The fine-tuned model with the new dataset generates improved results, yielding highly realistic and visually appealing pattern images..

## 2. EXPERIMENTAL STUDY

### 2.1 Diffusion Models

In this study, we use diffusion model architecture as a method for textile pattern generation. To improve the model's performance for this specific task, we fine-tune the pre-trained Stable Diffusion model with a custom dataset of textile patterns.

Denoising Diffusion models, such as the Stable Diffusion model, are inspired by considerations from nonequilibrium thermodynamics. Stable Diffusion differs from other proposed diffusion models in that it operates on the latent space rather than the pixel space, which reduces computational complexity and makes it more practical to implement with systems with limited computational power.

Overall, the Stable Diffusion model has demonstrated that diffusion models can be effectively utilized for generating high-quality images in the latent space domain and offers a promising approach for tasks such as text-guided textile pattern generation.

### 2.2 Dataset

In order to train our model for text-guided textile pattern generation, we have compiled a diverse dataset of approximately 4.000 pattern images from various sources on the internet. Some examples are given in Figure 1. Our dataset consists of a wide range of styles and designs to ensure that the model can generalize to a wide range of textile patterns. As a result of the promising initial results obtained from our model, we have also taken steps to improve further and expand our dataset to achieve even better performance.

To provide natural language supervision for our dataset, we have utilized a pre-trained image caption model called BLIP. This model has been trained on a large dataset of images and their corresponding captions and can generate textual descriptions of images based on their visual content. By using BLIP to generate captions for our dataset, we have provided our model with a rich source of textual information to guide the generation process. In addition to the captions generated by BLIP, we have also included a selection of keywords related to the style and content of the patterns in our dataset as an additional source of guidance for the model.

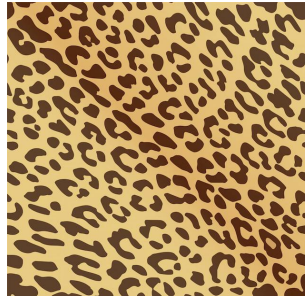 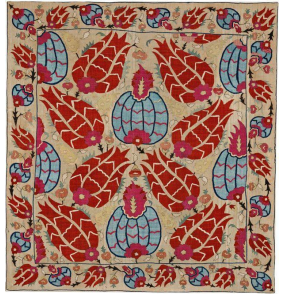 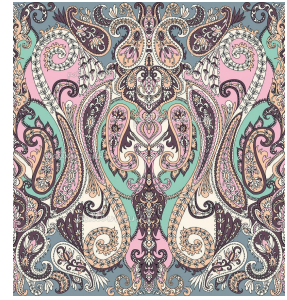

(a) Animal Patterns    (b) Ottoman Embroidery    (d) Indian Patterns

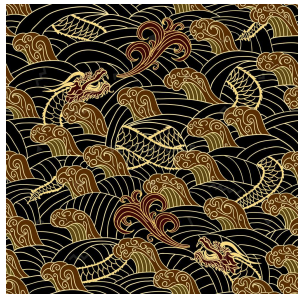 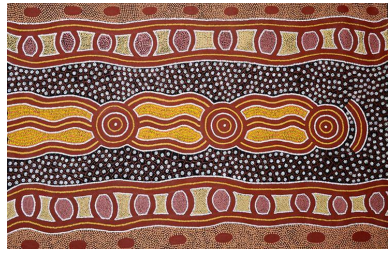 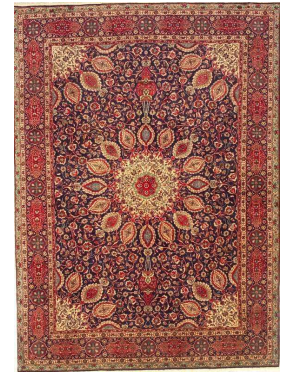

(c) Traditional Japanese    (e) Aboriginal Patterns    (f) Persian patterns

**Figure 1.** Samples from datasets and their keywords

**Table 1.** All keywords used to create the dataset.

| All Keywords In Our Dataset | | |
|---|---|---|
| Allover patterns | Abstract Patterns | Calico Patterns |
| Animal Patterns Printed | Ottoman Embroidery Patterns | Indian Fabric Patterns |
| Traditional Japanese Patterns | African Pattern Fabric | Iranian Rug pattern |
| Floral Pattern Fabric | Turkish Patterns | Oriental patterns |
| Aborigin patterns | Vintage Fabric Patterns | Art Nouveau pattern |

## 3. RESULTS

Our newly developed textile-generating diffusion model outperforms the original Stable Diffusion model. We reach this conclusion through both qualitative and quantitative evaluations.

We perform a qualitative evaluation of the results of the newly developed textile-generating diffusion model to compare its performance with the original Stable Diffusion model. The comparison can be seen in Table 2, where it is evident that the fine-tuned Textile Diffusion model outperforms the original Diffusion model across various textile patterns in given captions.

| Prompt | Our Model | Base Model |
| --- | --- | --- |
| Aboriginal Flower Pattern | | |
| Vintage Floral Patterns | | |
| Oriental Japanese Pattern | | |
| Turkish Patterns | | |
| Ottoman Embroidery Patterns | | |

**Table 2.** Qualitative comparison between the base model and proposed model

In addition to the qualitative evaluation, we use perceptual loss [6] and clip similarity for the quantitative evaluation. We calculate the clip similarity by analyzing keywords in the dataset given in Table 1 and using the CLIP model to compute the similarity between the generated images and the keywords. We calculate the perceptual loss by selecting base images randomly from the dataset related to the selected keywords and determining the differences in style and features between the generated images and the base images. It is seen in Table 3 that these results provide further support for the conclusion that the fine-tuned Textile Diffusion model outperforms the original Diffusion model.

|  | *Perceptual Loss (Lower is Better)* | | *Clip Similarity (Higher is Better)* | |
| :---: | :---: | :---: | :---: | :---: |
| **Keywords** | **Our Model** | Base Model | **Our Model** | Base Model |
| *Calico Pattern* | **30.13** | 71.08 | **65.31** | 57.0 |
| *Traditional Japanese Patterns* | **67.042** | 175.88 | **82.75** | 72.0 |
| *African Pattern fabric* | **132.26** | 164.88 | **79.87** | 73.31 |
| *Turkish Patterns* | **40.71** | 44.79 | **74.12** | 68.5 |
| *Oriental Patterns* | **49.70** | 115.76 | **78** | 73.87 |
| *Aboriginal Patterns* | **165.04** | 189.88 | **86.25** | 75.37 |
| *Modern Flower Print Fabric* | **80.13** | 101.78 | **74.75** | 73.18 |
| *Native American Pattern* | **66.19** | 84.57 | **81.0** | 70.56 |
| *Iranian Pattern* | **41.91** | 47.56 | **79.0** | 77.31 |

**Table 3.** Quantitative comparison between the base model and proposed model.

## 4. CONCLUSION

In conclusion, the fine-tuned Diffusion model represents a significant advancement in the field of text-guided textile pattern generation, providing improved results compared to the original model. This work has the potential to transform the way textile patterns are designed and created, streamlining the process and enabling the generation of intricate and sophisticated patterns with ease. The use of the Diffusion model for this specific task demonstrates its versatility and adaptability, making it a valuable tool for image generation in a wide range of domains.

Overall, this research contributes to the ongoing efforts in image generation, adding to the growing body of knowledge and expanding the capabilities of AI in this area. The ability to generate high-quality textile patterns based on text descriptions can revolutionize the textile industry, enabling designers to quickly and easily create complex patterns and bring new ideas to life. This work opens up exciting possibilities for the future of text-guided image generation, and we look forward to seeing the impact it will have in the field.